%% file: paper.tex
\documentclass[letterpaper, 10 pt, conference]{ieeeconf}  
\usepackage{float}

\IEEEoverridecommandlockouts                              

\usepackage{amsmath}
\usepackage{amssymb}
\usepackage{epsfig}
\usepackage{graphicx}
\usepackage{xcolor}
\usepackage{todonotes}
\let\labelindent\relax
\usepackage{enumitem}
\usepackage{enumitem}
\usepackage{stfloats}
\usepackage{tabularx,booktabs}
\newcolumntype{Y}{>{\centering\arraybackslash}X}
\usepackage{multirow}
\setlist[itemize]{leftmargin=*}

\title{\LARGE \bf
Built Different: Tactile Perception to Overcome Cross-Embodiment Capability Differences in Collaborative Manipulation
}
\newcommand{\mat}[1]{\mathbf{#1}}
\renewcommand{\vec}[1]{\boldsymbol{#1}}
\newcommand{\DI}{t}

\newcommand{\se}[1]{\mathrm{SE}(#1)}      

\newcommand{\proprio}{\vec{x}}

\newcommand{\shearfield}{\boldsymbol{\mathcal{U}}}

\newcommand{\wrench}{\vec{f}}

\newcommand{\horizontal}{\text{x}}
\newcommand{\vertical}{\text{y}}
\newcommand{\sol}{\text{sol}}
\newcommand{\irr}{\text{irr}}

\newcommand{\image}{\mat{I}}
\newcommand{\optflowfunc}{\mathbf{Flow}}

\newcommand{\vel}{\boldsymbol{v}}
\newcommand{\demopol}{\pi_D}
\newcommand{\studentpol}{\pi_S}
\newcommand{\block}[1]{\vspace{6pt}\noindent\textbf{#1.}}

\author{William van den Bogert$^{*1}$, Madhavan Iyengar$^{*2}$, Nima Fazeli$^{3}$
\thanks{*Equal contribution}
\thanks{$^{1}$ William van den Bogert is with the Mechanical Engineering Department at the University of Michigan, MI, USA 
{\tt\small willvdb@umich.edu}}%
\thanks{$^{2}$ Madhavan Iyengar is with the Computer Science Department at the University of Michigan, MI, USA 
{\tt\small miyen@umich.edu}}%
\thanks{$^{3}$ Nima Fazeli is with the Robotics Department at the University of Michigan, MI, USA
{\tt\small nfz@umich.edu}}
}

\begin{document}

\maketitle
\thispagestyle{empty}
\pagestyle{empty}

\begin{abstract}
    Tactile sensing is a widely-studied means of implicit communication between robot and human. In this paper, we investigate how tactile sensing can help bridge differences between robotic embodiments in the context of collaborative manipulation. For a robot, learning and executing force-rich collaboration require compliance to human interaction. While compliance is often achieved with admittance control, many commercial robots lack the  joint torque monitoring needed for such control. To address this challenge, we present an approach that uses tactile sensors and behavior cloning to transfer policies from robots with these capabilities to those without. We train a single policy that demonstrates positive transfer across embodiments, including robots without torque sensing. We demonstrate this positive transfer on four different tactile-enabled embodiments using the same policy trained on force-controlled robot data. Across multiple proposed metrics, the best performance came from a decomposed tactile shear-field representation combined with a pre-trained encoder, which improved success rates over alternative representations.
\end{abstract}

\input{text/01-introduction}

\input{text/02-related-work}

\input{text/03-methods}

\input{text/04-experiments}

\input{text/05-results}

\input{text/06_DL}


\bibliographystyle{ieeetr}
\bibliography{references}

\end{document}

%% file: text/01-introduction.tex
\section{Introduction}

Not all robots are built equal. Consider tasks that require force-rich physical interaction such as collaborative object carrying or object hand-over. High-end robotic platforms achieve these skills via impedance control and joint torque monitoring that enable the robot to balance position tracking and force exertion. However, a large number of industry-standard robot models produced by FANUC, ABB, and Kawasaki (among others) offer only position control with no access to joint torques. There are also affordable robotic platforms where the lack of dedicated joint-torque sensing is key for affordability. The central question that we address in this work is: ``How can we train a force-rich compliant policy on high-end robot data, and then transfer such a policy to robots equipped only with inexpensive tactile feedback?''

An alternative approach to addressing the difference in capability between embodiments is to mount auxiliary force-torque sensing at the robot ``wrist'' and use the resulting signal as a proxy for joint torque measurements. However, this approach has a number of important issues that are remedied by tactile sensing. First, the sensor readings are contaminated by inertial artifacts from robot acceleration, but tactile sensors are robust to this movement. Second, high-quality force/torque sensors are expensive (often thousands of dollars) and fragile, whereas the open-source tactile sensors used in this work cost only about \$130 in materials \cite{gelslim4} and provide inherent passive compliance for contact-rich interaction. Third, the sensors can occupy a significant portion of the robot's payload budget, whereas modern vision-based tactile sensors comprise of lightweight silicone pads. Fourth, summarizing an interaction with a single force vector may alias pertinent details of the task, but tactile sensing provides far richer feedback, enabling slip detection \cite{zha2018}, in-hand pose estimation, and extrinsic contact estimation \cite{kim2021}. There are other potential alternatives to both tactile sensing and auxiliary force-torque sensors, i.e. force-sensing resistors or primitive strain gauges. We argue that such alternatives require additional custom hardware, whereas modern research robots increasingly integrate tactile sensors, as reflected in recent literature \cite{don2021, gelslim4, auc2025, li2023}.

\begin{figure}[!t]
\centering
\includegraphics[width=1.0\columnwidth]{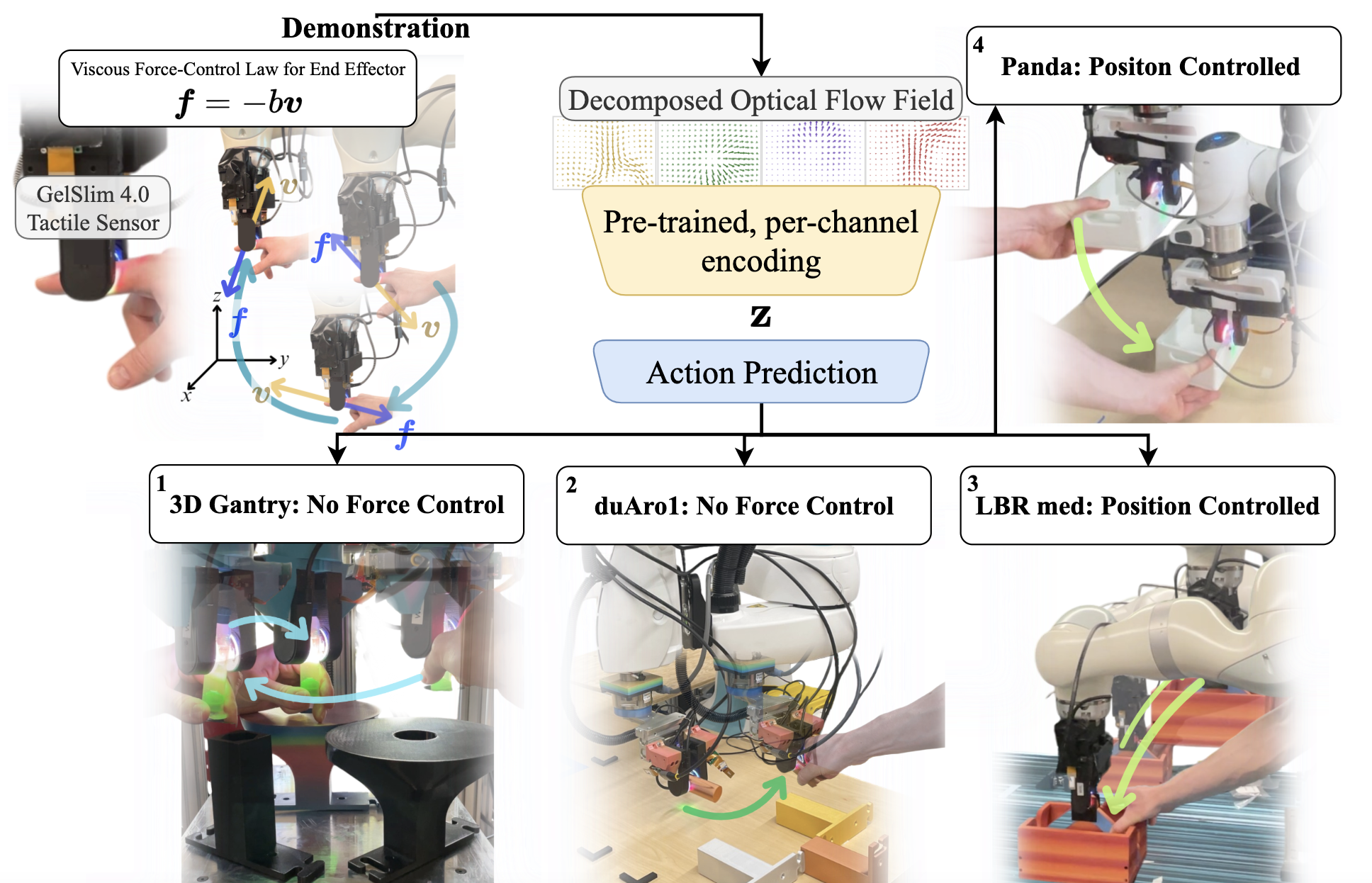}
\caption{Our framework for training a non-compliant robot to mimic compliant behavior during a collaborative task. The demonstration robot uses force feedback to enable a human to demonstrate the maneuvering of the grasped object throughout the workspace while tactile data is collected. A policy is trained from these demonstrations, then then rolled out four unique, purely position controlled embodiments.}
\label{fig:teaser}
\end{figure}

\subsection{Contributions}
This paper considers tactile sensing as a means of bridging embodiment differences to enable collaborative manipulation across robotic systems. We present a method that learns compliant behavior on an impedance-capable, torque-sensing robot, then transfers a single tactile policy to tactile-equipped robots without access to impedance control and joint-torque feedback.
Our key contributions are:
\begin{enumerate}
    \item \textbf{Tactile sensing as a practical proxy for force feedback} We show that inexpensive tactile sensors can provide sufficient feedback to enable compliant behavior typically achieved via impedance control, even on robots without torque sensing.
    \item \textbf{Structured tactile representations for policy transfer.} We demonstrate that tactile representations derived from shear displacement fields, and particularly their Helmholtz-Hodge decompositions, improve data efficiency and generalization in learning our compliant policy when we test across four distinct embodiments.
\end{enumerate}

\subsection{Problem Statement} \label{problemstatement2}

We consider two robotic systems:
(1) a high-end \emph{demonstration} robot with joint-torque sensing and impedance control, and
(2) a \emph{student} robot without torque sensing or impedance control.
Our goal is to learn a policy from demonstrations on the high-end robot that enables the student robot, when equipped with tactile sensing, to exhibit similar compliant behavior during collaborative manipulation. 

\block{Workspace Assumption} We assume Cartesian workspace of the student robot $\mathcal{W}_S$ is contained within that of the demonstration robot $\mathcal{W}_D$:
\begin{equation*}
    \mathcal{W}_S \subseteq \mathcal{W}_D \in \se 3.
\end{equation*}
This ensures demonstrations can be collected within the student robot’s reachable space.

\block{Demonstration Robot Control} On the demonstration robot, we implement an impedance control law $g$, mapping end-effector velocity $\vel\in\mathbb{R}^{\text{dim}(\mathcal{W}_S)}$ and position $\proprio\in\mathcal{W}_S$ to wrench $\wrench\in\mathbb{R}^{\text{dim}(\mathcal{W}_S)}$:
\begin{equation*}
    \wrench = g(\vel,\proprio)\label{eq:elastic}
\end{equation*}
For our compliance problem, we focus on a purely viscous control regime, assuming $g$ depends only on velocity:
\begin{equation}
    \wrench = g(\vel)\label{eq:viscous}
\end{equation}
Assuming $g$ is invertible, we express the demonstration robot's compliant policy $\demopol:\mathbb{R}^{\text{dim}(\mathcal{W}_S)}\rightarrow\mathbb{R}^{\text{dim}(\mathcal{W}_S)}$ as:
\begin{equation}
    \vel = \demopol(\wrench)\label{eq:demopol}
\end{equation}
where $\demopol$ maps sensed forces to end-effector velocities.

\block{Relating Tactile Feedback to Forces} Let the vector field $\mathbf{u} \in V_{\mathbb{R}}$ represent the continuous deformation of a vision-based tactile sensor. We approximate this discretely via a shear-displacement tensor $\shearfield \in \mathbb{R}^{C \times H \times W}$, where $C$ channels are defined over an $H \times W$ grid. Because the deformable material of the sensor exhibits roughly visco-elastic properties, we assume a time-invariant mapping $\mathcal{M}:V_{\mathbb{R}}\rightarrow\mathbb{R}^{\text{dim}(\mathcal{W}_S)}$ of tactile feedback to end-effector forces:
\begin{equation*}
    \wrench = \mathcal{M}(\boldsymbol{u}) \approx \mathcal{M}'(\shearfield)\label{eq:mapping}
\end{equation*}
where $\mathcal{M}':\mathbb{R}^{C\times H \times W}\rightarrow\mathbb{R}^{\text{dim}(\mathcal{W}_S)}$ is an approximation which is learnable from data \cite{zha2019, zha2022, auc2025}.

\block{Student Policy via Mapping Composition} Substituting the tactile-to-force mapping into Eq.~\ref{eq:demopol} yields:
\begin{equation*}
    \vel \approx \demopol(\mathcal{M}'(\shearfield)) \label{eq:plugin}
\end{equation*}
We define the student policy $\studentpol$ as the composition of the force-to-motion policy $\demopol$ with the tactile-to-force map $\mathcal{M}'$:
\begin{equation}
    \vel \approx \studentpol(\shearfield) \label{eq:student}
\end{equation}
where $\studentpol = \demopol \circ \mathcal{M}'$.

By approximating $\studentpol$ via supervised imitation learning of a demonstration dataset of a tactile-equipped \emph{demonstration robot}'s behavior, a tactile-equipped \emph{student robot} can perform $\studentpol$ without impedance-control or related capabilities.

This formulation highlights why vision-based tactile sensing can serve as a practical proxy for force feedback, enabling compliant policy transfer across embodiments lacking torque sensing. We implement this formulation with tactile sensing fingertips, though  it can be applied to other tactile anatomies (i.e. robot skin) as well.

Transferring our policy across embodiments presents a series of challenges. Policy success on unseen robotic systems is inhibited by the diverse set of properties (base stiffness, grasp strength, control architecture, environmental conditions, grasped objects) exhibited by embodiments. We aim to overcome these challenges by selecting a structured representation for the shear-displacement tensor $\shearfield$ based on the Helmholtz-Hodge decomposition.

\begin{figure}[t]
\centerline{\includegraphics[width=0.76\columnwidth]{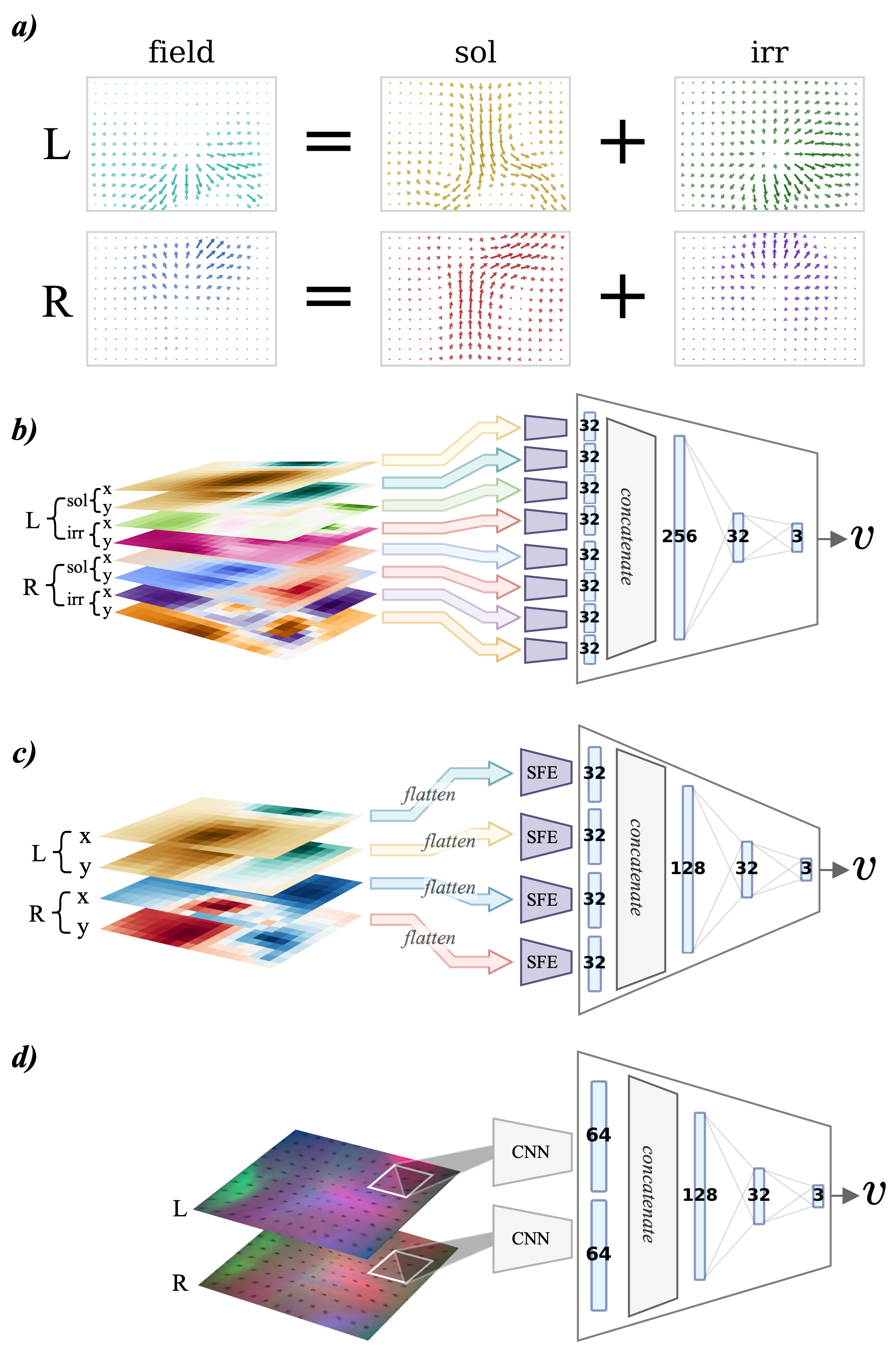}}
\caption{(a) Example of Helmholtz-Hodge Decomposition (HHD). (b-d) The various tactile representations and their associated neural network architectures: (b) HHD, (c) shear-field, and (d) tactile image input to our model.}
\label{fig:arch}
\end{figure}
\vspace*{-2mm}

%% file: text/02-related-work.tex
\section{Related Works}
\subsection{Tactile Perception}
Tactile information is critical for humans and robots alike. Studies of human tactile perception demonstrate its role in force estimation and fine motor control \cite{jon2014, gor2004}. In robotics, tactile sensing has long been used as an input for wrench estimation \cite{zhou2024}, including with GelSight \cite{yuan2017, lu2024} and modified GelSight \cite{li2023} sensors. To the best of our knowledge, the role of tactile sensing in compliant robotics remains underexplored, but there are works in this research domain. \cite{fath2007, burgess2024learningobjectcomplianceyoungs} used tactile sensing to detect object compliance. \cite{jentoft2014, Yang2024} demonstrated compliant grasping control with tactile-equipped end-effectors, as a means to improve grasp safety and quality. However, these works are limited to end-effector compliance or force perception, without extension to whole-robot compliant behavior or cross-embodiment collaborative task transfer, for which we learn a policy in this work.

\subsection{Tactile Policy Learning}
Whole-robot tactile policies have been learned in both reinforcement learning (RL) and behavior cloning (BC) frameworks. For a robot tasked with peg-in-hole insertion, \cite{lee2020} shows that an RL policy succeeds most frequently with multimodal vision and tactile (in the form of force-torque) feedback. These modalities enable deformable object manipulation policies as well \cite{wi2022virdorealworldvisuotactiledynamics, vandermerwe2023integratedobjectdeformationcontact}.
Similar RL work was later implemented by \cite{don2021} with vision-based tactile sensors, indicating that a force-rich task can succeed with just fingertip tactile feedback.

Our work focuses on tactile policies generated from BC. Behavior cloning state-of-the-art Diffusion Policy \cite{chi2023} has been integrated with tactile feedback before \cite{lin2024,geo2024}. Diffusion Policy is well-suited to intelligently generate trajectories of end-effector poses from multimodal feedback. Our work requires a more streamlined architecture, in which a single-step velocity action is generated in response to tactile feedback. This choice reflects our goal of cloning the behavior of the demonstration robot's viscous control regime with only vision-based tactile sensing--no impedance/admittance control, force-sensing resistors, joint current/torque monitoring, or other force sensing solutions.

\subsection{Structured Representations}
We use the GelSlim 4.0 vision-based tactile sensor, which captures high-dimensional deformation signals, but produces state distributions sensitive to lighting and grasp variability. To improve data efficiency, previous work has distilled raw tactile images into structured representations, such as a discrete field of shear displacements \cite{zha2018}. This field can be decomposed using Helmholtz-Hodge decomposition (HHD) \cite{hel1858}, which has precedent in tactile sensing. \cite{zha2023} uses HHD as a constraint to learn a representation of the tactile shear field. In contrast, our work uses HHD to decompose the shear field obtained from optical flow measurements. \cite{zha2022}, \cite{auc2025}, and \cite{zha2019} use a similar method to perceive a contact wrench. Our work bypasses the force estimation step and learns a representation of the HHD to enable our collaborative policy. Since this representation mitigates sensitivity to lighting and grasp variability, the same collaborative policy works across tactile-equipped embodiments.

\subsection{Cross-Embodiment Transfer}
Prior work has explored cross-embodiment policy transfer using latent space alignment \cite{wang2024crossembodimentrobotmanipulationskill} and raw sensory alignment \cite{niu2024}. Transfer of vision-based policies has also been improved by in-painting the demonstration robot into the student robot’s scene \cite{mirage}.  

In contrast, we enable zero-shot transfer of a collaborative tactile policy by using the learned HHD representation as a shared state space between the demonstration and student robots. Importantly, this approach does not require any additional data collection on the student robot before transfer.

%% file: text/03-methods.tex
\section{Methods}\label{section:methods}
Our framework is intended to mimic the compliant behavior of impedance-controlled robots using BC and a structured tactile representation.

\begin{figure*}[t]
\vspace{2mm}
\centerline{\includegraphics[width=0.95\textwidth]{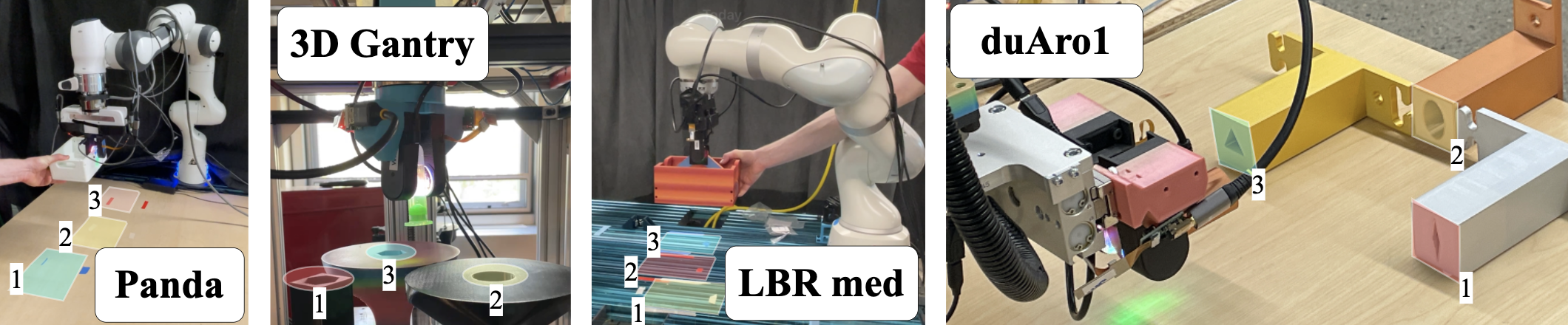}}
\caption{Policies on these embodiments were evaluated on the task of maneuvering grasped objects to the three highlighted goal locations.}
\label{fig:goal_loc}
\end{figure*}

\begin{figure}[b]
\centerline{\includegraphics[width=0.9\columnwidth]{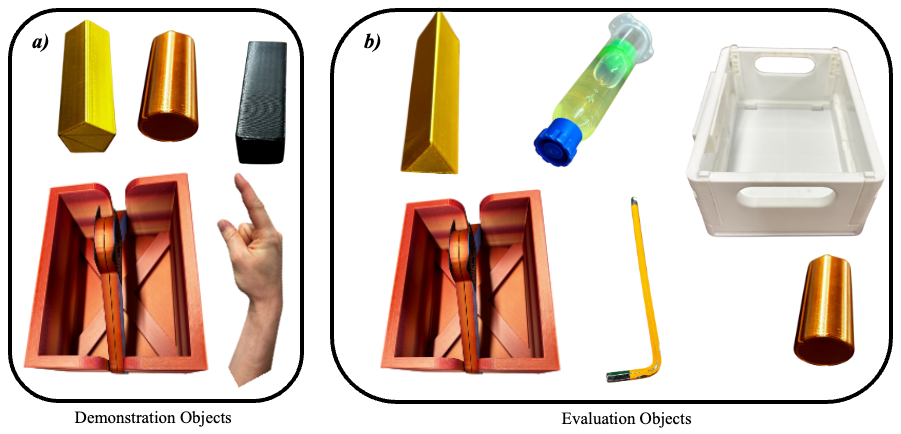}}
\caption{Objects grasped by the robot during (a) demonstration collection and (b) evaluation.}
\label{fig:object}
\end{figure}

We focus on manipulators with parallel-jaw grippers where end-effector forces $\wrench$ arise from human interaction with a grasped object. While our method uses fingertip tactile sensing, it could also extend naturally to other tactile modalities (e.g., robot skin), though we leave this to future work.

\subsection{Tactile Representations}\label{tac}
To learn a tactile-informed policy that (1) mimics compliant behavior and (2) generalizes well across embodiments, we require a structured representation of sensor feedback.

The GelSlim 4.0 sensor \cite{gelslim4} we use in this work enables high-resolution tactile sensing in the form of raw RGB images $\image$. To derive the shear-displacement tensor from these images, we used the \textbf{open-cv2} Python library. A function $\optflowfunc$ calculates the optical flow components from the undeformed image $\image_0$ to the deformed image $\image_\DI$. We then use the Python library \textbf{naturalHHD} \cite{bha2014} to estimate the solenoidal ($\sol$) and irrotational ($\irr$) components of the Helmholtz-Hodge decomposition (HHD) \cite{hel1858}, where:
\begin{equation*}
\sol + \irr = (\horizontal, \vertical) = \optflowfunc(\image_0,\image_\DI)\label{eq:optical_flow}
\end{equation*}
where x and y are the horizontal and vertical components of the vectors, respectively. This decomposition, visualized in Fig.~\ref{fig:arch}a, separates divergence-free and curl-free motion patterns, which we hypothesize as decoupling normal and shear force information further than raw optical flow.

For a parallel-jaw gripper with two GelSlim sensors, we obtain two decompositions (L, R), each with two fields and two components, resulting in an 8-channel tensor $\shearfield \in \mathbb{R}^{8\times 13 \times 18}$ (see Section~\ref{problemstatement2}).

\subsection{Scalar Field Encoder}\label{SFE}
Each channel of $\shearfield$ is processed by a scalar field encoder (SFE), shown in Fig.~\ref{fig:arch}b. This fully connected network is pre-trained on reconstruction of channels from a tactile manipulation dataset. The encoder outputs an embedding $\mathbf{z}_{i} \in \mathbb{R}^{32} = \text{SFE}(\shearfield_{i}) \text{ for } i\in\{0\cdots7\}$. Fig. \ref{fig:arch}b shows the reconcatenation of these fields post-encoding, providing a latent representation of the entire HHD $\mathbf{z_{\shearfield}}\in\mathbb{R}^{256}$.

\subsection{Data Collection of Compliant Behavior} \label{subsection:supplant}
To collect demonstrations, we exploit the impedance-control capabilities of the demonstration robot to generate compliant behaviors. Following Eq.~\ref{eq:viscous}, we implement a viscous control regime where the end-effector exerts a wrench $\wrench \in \mathbb{R}^3$ opposite to its velocity $\vel \in \mathbb{R}^3$:
\begin{equation}
    \wrench = -\mathbf{B}\vel\label{eq:linear_law}
\end{equation}
where $\mathbf{B} > 0$ is a diagonal damping matrix. Intuitively, this models the robot as resisting motion proportionally to velocity, allowing it to ``yield” naturally to human guidance. With this control active, we collect the synchronized dataset $\mathcal{D} = \{\shearfield, \vel\}$ from the tactile-equipped demonstration robot.

\subsection{Behavior Cloning Architecture}\label{subsection:MLP}
Given $\mathcal{D}$, we solve a supervised imitation learning problem to find $\studentpol:\mathbb{R}^{C\times H\times W}\rightarrow\mathbb{R}^3$ from Eq. \ref{eq:student}.

We use a lightweight 3-layer multilayer perceptron (MLP) architecture, shown in Fig. \ref{fig:arch}, to map the embedding of tactile state $\mathbf{z_{\shearfield}}$ to velocity $\boldsymbol{v}$:
\begin{equation}
    \boldsymbol{v}\approx\text{MLP}(\mathbf{z_{\shearfield}})
\end{equation}
We train this MLP on L2-norm regression to $\vel$ given paired $\mathbf{z_{\shearfield}}$ in the dataset $\mathcal{D}$. The combination of the MLP and the concatenated SFE outputs is our estimate for $\studentpol$:
\begin{equation}
    \vel\approx\studentpol(\shearfield)\approx\text{MLP}(\text{SFE}(\shearfield_0)\oplus\cdots\oplus\text{SFE}(\shearfield_7))
\end{equation}
After training, $\studentpol$ predicts the velocity that imitates $\demopol$.

%% file: text/04-experiments.tex
\section{Experiments}\label{section:experiments}
In this section we describe our experimental setup, baseline methods, and evaluation metrics. With our experiments, we aim to show that
(1) compliant behavior can in fact be enabled entirely through vision-based tactile feedback, serving as an effective proxy for joint torque sensing.
(2) the compliant behavior we learn generalizes across diverse robot embodiments and grasped objects.
(3) this generalization is best achieved through structured tactile representations, such as the HHD proposed in Section \ref{tac}.

\subsection{Training}
\block{Data Collection} Training demonstrations were collected on a 7-DOF Kuka LBR Med manipulator constrained to a 3D translational workspace. The demonstration robot executed the viscous control regime from Eq.~\ref{eq:linear_law}, while grasping one of the objects shown in Fig.~\ref{fig:object}a. A human operator manipulated the grasped object across the workspace, aiming to generate a wide range of velocities in both magnitude and direction. We collected 62 episodes, each using a different object, resulting in roughly $27,\!000$ $(\shearfield, \vel)$ pairs.

\block{Dataset Filtering} Since many samples contained near-zero velocities, which can bias behavior cloning toward static actions, we filtered out data points with $||\vel|| < 2\times\mathrm{median}(||\vel||)$, resulting in approximately 3700 data points.

\block{Data Augmentation} We augmented the filtered dataset by adding Gaussian noise ($\sigma=0.1$) to $\shearfield$, expanding the dataset to roughly 15,000 usable samples.

\block{Dataset Split} We used an 80–10–10 train–validation–test split such that around 12,000 data points were seen by the network during training.

\begin{figure}[t]
\vspace{2mm}
\centerline{\includegraphics[width=0.86\columnwidth]{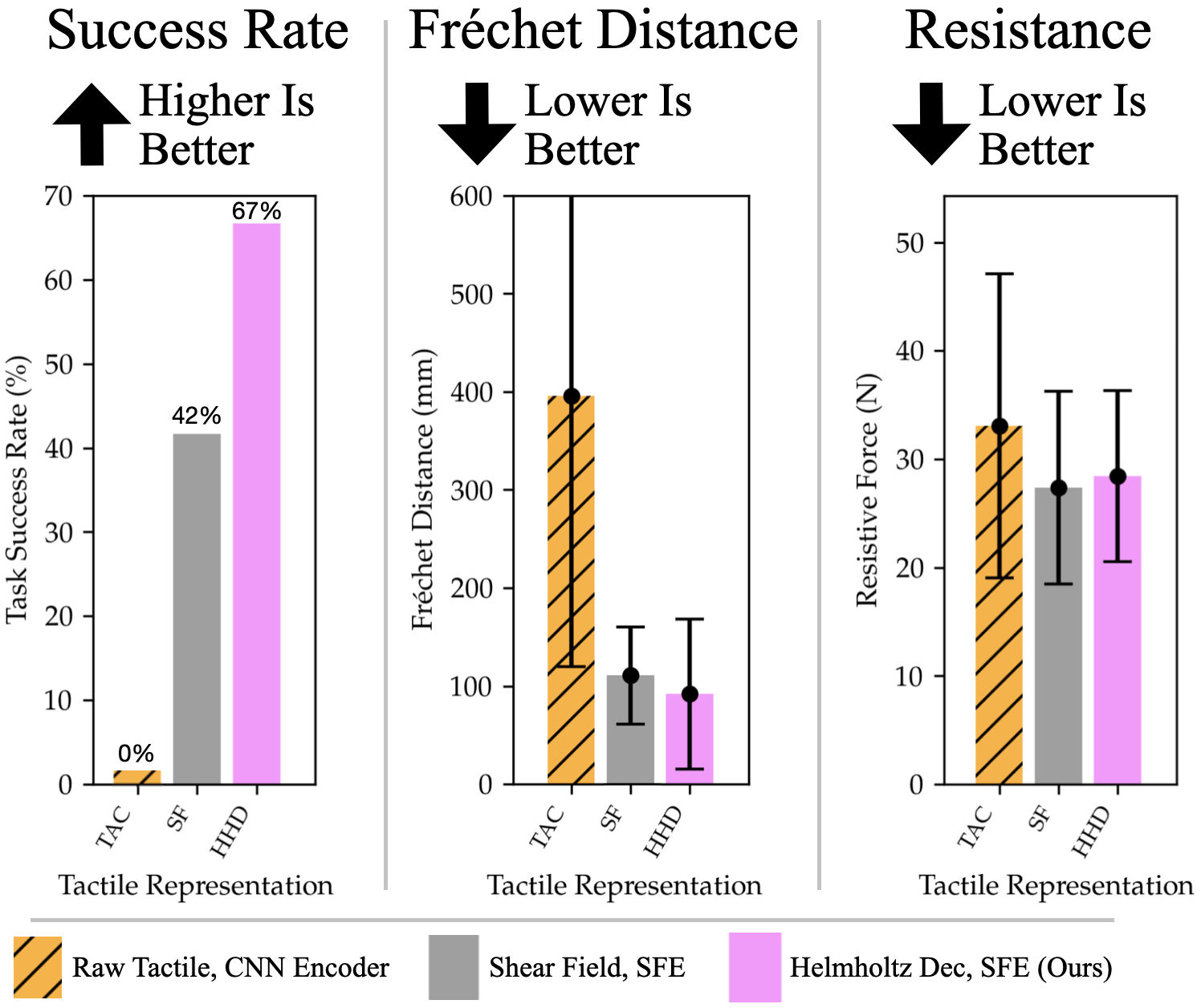}}
\caption{Focusing on policies with pre-trained encoders only, this graph shows the effect of different tactile representations on compliant behavior over all test embodiments. Performance is improved when HHD is used over other representations. Error bars represent $1\sigma$.}
\label{fig:representation}
\end{figure}

\subsection{Evaluation}\label{section:eval}

\subsubsection{Student Robot Embodiments}
We evaluated trained policies on 4 student robots (Fig.~\ref{fig:goal_loc}), each differing in morphology and control capabilities:
\begin{enumerate}
    \item{\textbf{Panda:} a small manipulator set to position-control mode, cut off from native force monitoring.}
    \item{\textbf{3D Gantry:} a 3D prismatic gantry adapted from a 3D printer, with no native force-control capabilities.}
    \item{\textbf{LBR Med:} the same manipulator as the demonstration robot, though cut off from native force monitoring.}
    \item{\textbf{duAro1:} a manipulator without native force monitoring.}
\end{enumerate}

\subsubsection{Goal-Reaching Tasks}
In each trial, a student robot grasped an object from Fig.~\ref{fig:object}b and the human operator guided the object toward 1 of 3 goal locations (Fig.~\ref{fig:goal_loc}). All 3 goals were tested independently for each embodiment.

\subsubsection{Metrics for Evaluation}\label{sec:metrics}
\begin{figure*}[t]
\vspace{1mm}
\centerline{\includegraphics[width=\textwidth]{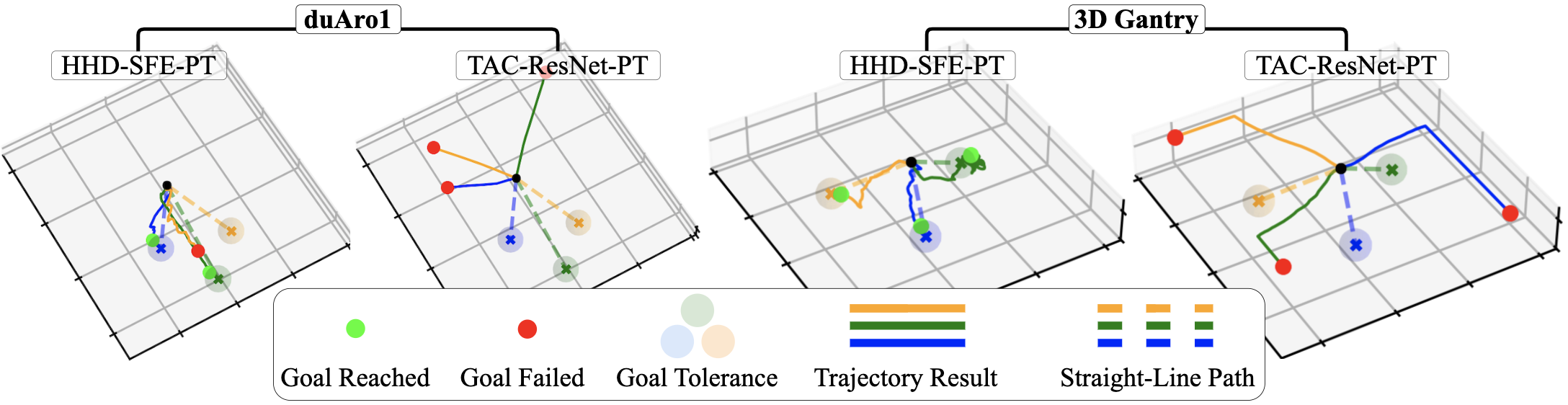}}
\caption{Example trajectories from our results.}
\label{fig:rollouts}
\end{figure*}

We evaluated our method on six different metrics in order to holistically determine which policies perform best.
\begin{enumerate}
    \item{\textbf{Fréchet distance from straight-line path} was computed between the executed trajectory and the straight-line path from the initial pose to the target, intending to capture global similarity in shape and ordering.}
    \item{\textbf{Projected root-mean-square error (RMSE)} was also computed between the executed trajectory and the straight-line path, intending to assess local deviations. As this is a measure of deviation from points sampled from the executed trajectory to their projection on the straight-line path, we only consider this metric to be valid if the goal-location is reached. Otherwise, a student robot which does not move gets a perfect score.}
    \item{\textbf{Path inefficiency} was defined as the ratio of the executed path length to the straight-line distance, where lower values indicate more efficient execution. This was also recorded for successful runs only. Otherwise, a student robot which does not move gets a perfect score.}
    \item{\textbf{Resistive force} was measured as the average magnitude of the end-effector force applied over the entire trajectory, serving as a proxy for human effort and recorded via an external ATI Gamma force-torque sensor.}
    \item{\textbf{Task duration:} was recorded for successful runs only, as another measure of efficiency.}
    \item{\textbf{Success rate:} the fraction of trials where the student robot's final position falls near the goal location (illustrated in Fig.~\ref{fig:goal_loc}), with a tolerance of 15\% (visualized in Fig. \ref{fig:rollouts}) of the workspace size (via longest dimension of 3D rectangular prism) of the student robot.}
\end{enumerate}
As these metrics are non-standard, we aim to make a recommendation of a subset of metrics based on our results.

\begin{figure}[b]
\centerline{\includegraphics[width=0.86\columnwidth]{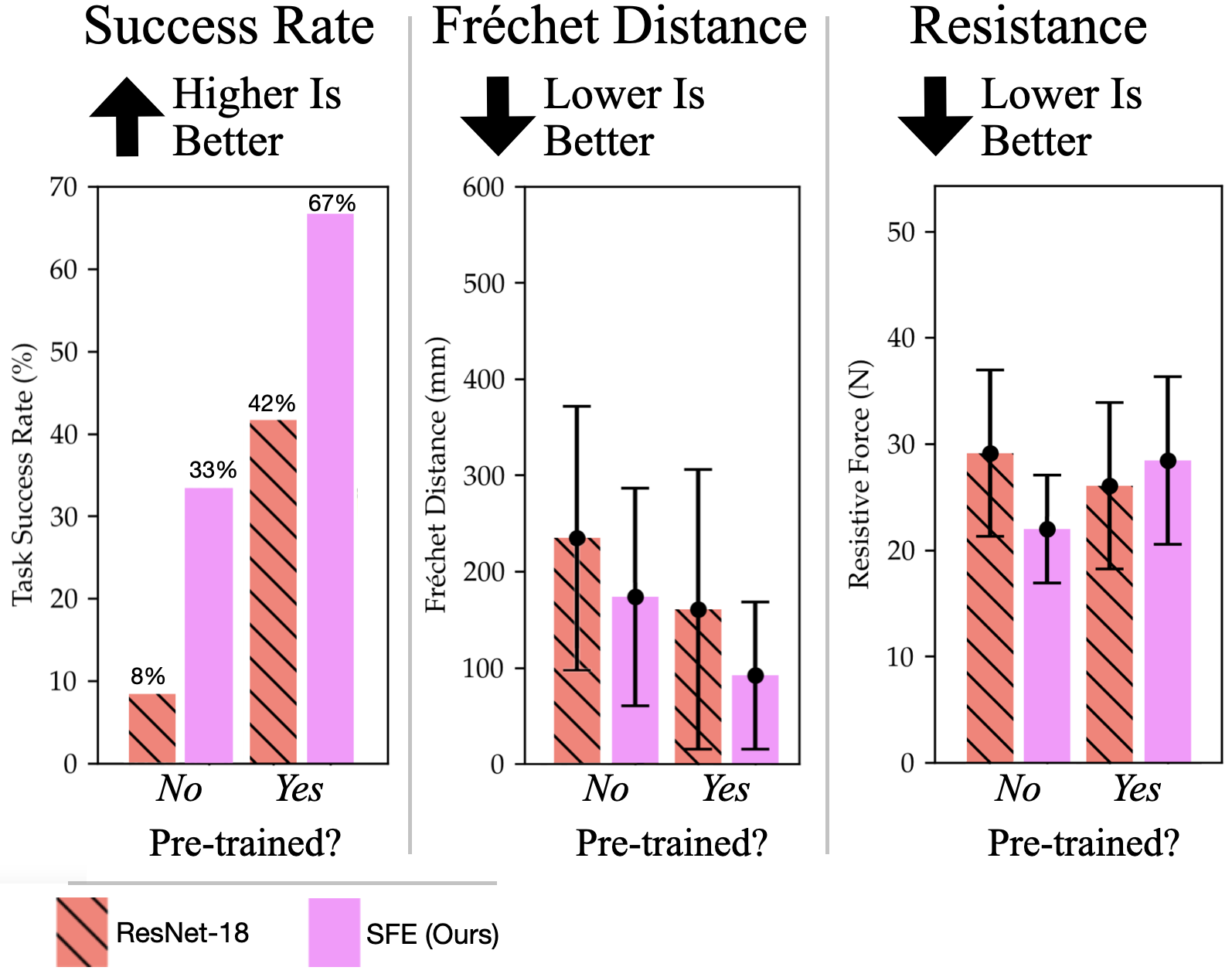}}
\caption{Focusing on HHD policies only, this graph shows the effect of different tactile encoders on compliant behavior over all test embodiments. Performance is improved our pre-trained SFE is used over other variants. Error bars represent $1\sigma$.}
\label{fig:pretraining}
\end{figure}

\subsection{Baselines}\label{section:baselines}
We evaluated a suite of model variants spanning different tactile representations, encoders, and training strategies in order to thoroughly baseline our proposed method.

\subsubsection{Tactile Representations}
We investigated 3 methods of representing tactile information prior to its presentation to any learning algorithm.
\begin{itemize}
\item{\textbf{Raw Tactile Image (TAC)} baselines test whether high-dimensional GelSlim images, without preprocessing, are sufficient for learning transferable collaborative behavior}
\item{\textbf{Shear-Field (SF)} baselines apply the optical flow function as discussed in Section \ref{tac}, without the HHD.}
\item{\textbf{Helmholtz-Hodge decomposition (HHD)} baselines apply the decomposition as discussed in Section \ref{tac}.}
\end{itemize}

\subsubsection{Encoders}
We investigated 2 encoder architectures for each tactile representation: a ResNet-18 and another using a neural network with a smaller number of parameters.
\begin{itemize}
\item{\textbf{ResNet-18 (ResNet)} is a commonly used residual learning encoder for vision \cite{kai2015}. We employ the 18-layer version.}
\item{\textit{Smaller Encoder \#1: }\textbf{3-layer Convolutional Neural Network (CNN)} is a custom low-dimensional convolutional neural network we apply to the TAC representation only, since this representation is too high dimensional for the fully-connected scalar field encoder (SFE).}
\item{\textit{Smaller Encoder \#2: } \textbf{Scalar Field Encoder (SFE)} is our proposed scalar field encoder discussed in Section \ref{SFE}, which we apply to both SF and HHD representations.}
\end{itemize}

\subsubsection{Training Strategies}
We investigated 2 training strategies for our behavior cloning architecture.
\begin{itemize}
\item{\textbf{Pre-training (PT)} involves training the above encoders on reconstruction of their given tactile representation, given a dataset with a larger distribution of tactile information than the imitation learning dataset $\mathcal{D}$. The pre-training dataset is collected from a variety of tactile manipulation scenarios.}
\item{\textbf{End-to-end training (E2E)} involves training the above encoders simultaneously with the MLP that predicts velocity actions, given only the imitation learning dataset $\mathcal{D}$.}
\end{itemize}

\noindent Combining $3$ representations, $2$ encoders, and $2$ training strategies yields $12$ model variants, including our proposed approach (HHD + SFE + PT). Each policy was trained on the same dataset $\mathcal{D}$ and evaluated on all 4 robot embodiments. Each of the 3 goal-reaching tasks used a model trained with 1 of 3 independent random seeds. Thus, each of the 12 model variants received 12 independent evaluations.

%% file: text/05-results.tex
\begin{table*}[bp]
\renewcommand{\arraystretch}{0.7}
\centering
\caption{Fréchet Distance From Straight-Line Path Results (mm) \textbf{↓ Lower is better.} | Mean $\pm$ Std. Deviation Over 3 Goal Locations}
\label{tab:frechet}
\begin{tabularx}{\textwidth}{r *{4}{Y}}
\toprule
\textbf{Method} & \textbf{LBR Med} & \textbf{Panda} & \textbf{3D Gantry} & \textbf{duAro1} \\
\midrule
\text{TAC-ResNet-E2E} & 989 $\pm$ 334 & 190 $\pm$ 66.7 & 286 $\pm$ 36.3 & 351 $\pm$ 127 \\
\text{TAC-CNN-E2E} & 521 $\pm$ 268 & 175 $\pm$ 27.8 & 212 $\pm$ 52.3 & 209 $\pm$ 76.5 \\
\text{TAC-ResNet-PT} & 922 $\pm$ 355 & 199 $\pm$ 54.6 & 268 $\pm$ 64.9 & 435 $\pm$ 153 \\
\text{TAC-CNN-PT} & 813 $\pm$ 197 & 166 $\pm$ 81.6 & 344 $\pm$ 59.8 & 260 $\pm$ 89.5 \\
\midrule
\text{SF-ResNet-E2E} & 563 $\pm$ 264 & 102 $\pm$ 23.5 & 273 $\pm$ 67.6 & 106 $\pm$ 27.2 \\
\text{SF-SFE-E2E} & 379 $\pm$ 237 & \textbf{35.0 $\pm$ 12.3} & 99.3 $\pm$ 52.2 & 87.5 $\pm$ 49.0 \\
\text{SF-ResNet-PT} & 186 $\pm$ 97.7 & 80.5 $\pm$ 75.1 & 194 $\pm$ 125 & 211 $\pm$ 73.5 \\
\text{SF-SFE-PT} & 132 $\pm$ 69.9 & 103 $\pm$ 57.4 & \textbf{91.1 $\pm$ 17.5} & 116 $\pm$ 17.6 \\
\midrule
\text{HHD-ResNet-E2E} & 360 $\pm$ 137 & 103 $\pm$ 63.7 & 257 $\pm$ 63.2 & 218 $\pm$ 118 \\
\text{HHD-SFE-E2E} & 267 $\pm$ 173 & 138 $\pm$ 14.5 & 133 $\pm$ 36.8 & 155 $\pm$ 88.4 \\
\text{HHD-ResNet-PT} & 295 $\pm$ 165 & 38.9 $\pm$ 9.26 & 247 $\pm$ 81.0 & 61.8 $\pm$ 22.7 \\
\text{HHD-SFE-PT} & \textbf{76.8 $\pm$ 10.9} & 38.4 $\pm$ 8.93 & 177 $\pm$ 106 & \textbf{74.4 $\pm$ 33.1} \\
\bottomrule
\end{tabularx}
\end{table*}

\section{Results}\label{section:results}

This section reports performance across all model variants and robot embodiments described in Section~\ref{section:experiments}.  
To simplify references, we abbreviate model variants as:
\begin{center}
\{Tactile Representation\}-\{Encoder\}-\{Training Strategy\}
\end{center}

\noindent where:
\begin{enumerate}
    \item \textbf{Tactile Representation} is either \textbf{HHD} (Helmholtz-Hodge decomposition), \textbf{TAC} (raw GelSlim images) or \textbf{SF} (non-decomposed shear-field).
    \item \textbf{Encoder} is either \textbf{SFE} (our scalar field encoder), \textbf{ResNet} (ResNet-18), or \textbf{CNN} (a 3-layer CNN).
    \item \textbf{Training Strategy} is either \textbf{PT} (pre-trained encoder), or \textbf{E2E} (simultaneous training of encoder and MLP)
\end{enumerate}

\noindent For example, HHD-SFE-PT is our proposed method. This section contains results for each model variant based on evaluations described in Section \ref{section:experiments} on the four different robot embodiments: Panda, 3D Gantry, LBR Med, and duAro1.

\subsection{Key Findings}
These results aggregate performance across all embodiments to isolate the improved effect of (1) the HHD representation over other choices and (2) pre-training the HHD SFE encoder over E2E and ResNet variants.

\subsubsection{Effect of Tactile Representation}
Fig. \ref{fig:representation} compares models using different tactile representations, all paired with their smallest viable pre-trained encoder (SFE for HHD and SF, CNN for TAC). Here we present the metrics which are valid for both successful and unsuccessful trials, with success defined as in Section \ref{section:eval}: success rate, Fréchet distance, and resistive force. HHD generates the highest success rate at 67\% over SF (42\%) and TAC (0\%). On Fréchet distance, HHD performs slightly better than SF on average, while TAC consistently underperforms. The resistive-force metric does not reveal strong trends across methods and is therefore less informative for evaluating policy quality. These results suggest that structured representations—particularly HHD—are better suited for this problem than raw GelSlim images.

\subsubsection{Effect of Encoder and Training Strategy}
Fig. \ref{fig:representation} isolates the HHD representation to compare encoders and training strategies. 
We again present the metrics which are valid for both successful and unsuccessful trials. Among the HHD-based models, our pre-trained SFE achieves the highest success rate ($67\%$) compared to E2E SFE ($42\%$) and ResNet-18 variants (pre-trained $33\%$, E2E $8\%$). Fréchet distance follows a similar trend, favoring the pre-trained SFE. Together, these findings indicate that lightweight, task-specific encoders benefit from pre-training on reconstruction tasks before behavior cloning, while larger generic encoders such as ResNet-18 appear less effective for this structured input.

\subsection{Comprehensive Results}
\begin{figure*}[h]
\centerline{\includegraphics[width=0.95\textwidth]{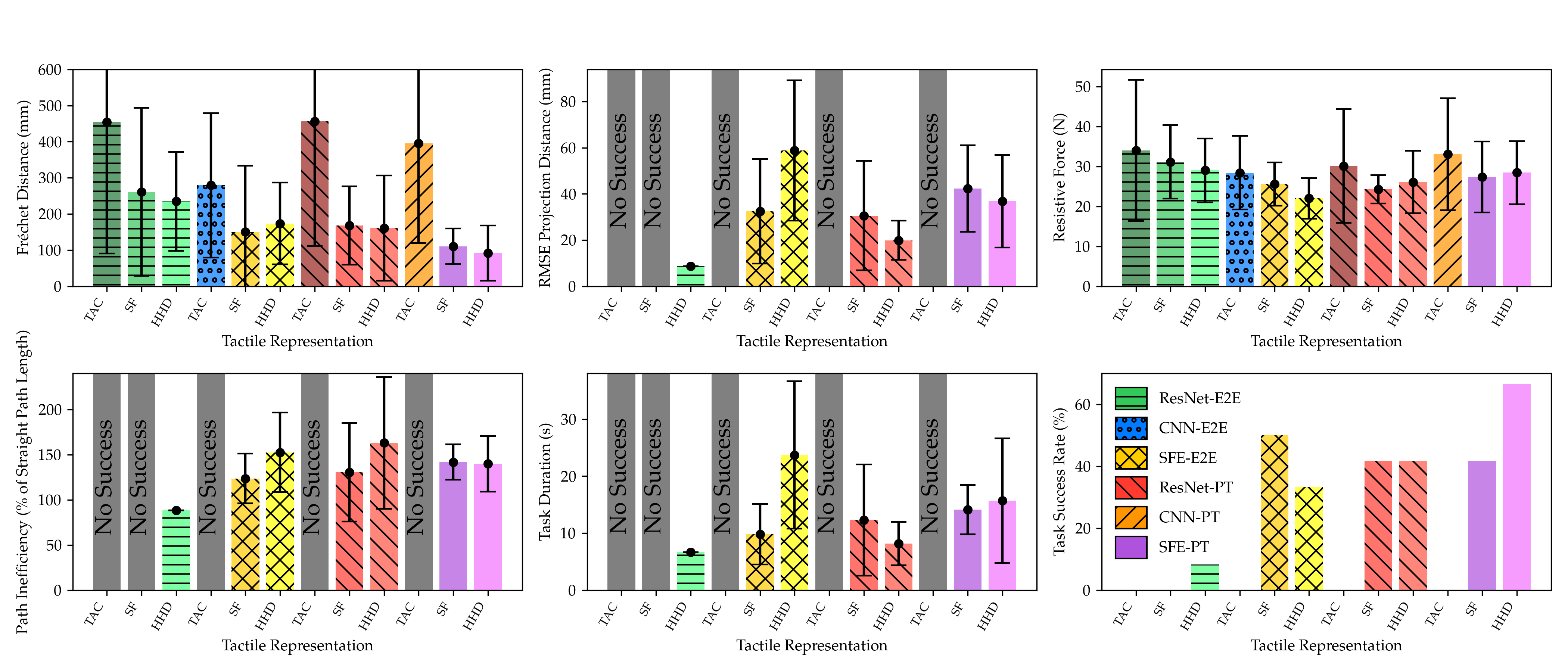}}
\caption{All metrics evaluated for each method over all rollouts on all embodiments. A gray ``No Success" bar indicates the lack of valid results given that there where no rollouts where the goal location was reached. Error bars represent $1\sigma$.}
\label{fig:bd_results}
\end{figure*}

Figure~\ref{fig:bd_results} summarizes the 6 evaluation metrics from Section~\ref{section:eval} across the 12 model variants, averaged over all embodiments. Among these, \textbf{Fréchet distance} and \textbf{success rate} emerge as the most informative, providing clear indications of trajectory quality and task completion, respectively.

In contrast, \textbf{resistive force} shows little variation between models and correlates weakly with other measures of performance, making it less useful for differentiating methods. Finally, the remaining metrics—\textbf{RMSE}, \textbf{path inefficiency}, and \textbf{task duration}—offer complementary insights but are less reliable when success rates are low, since averages are then based on fewer trials.

\begin{table*}[bp]
\centering
\caption{Task Success Rate Results (\%) \textbf{↑ Higher is better} | Mean Over 3 Goal Locations}
\label{tab:success}
\begin{tabularx}{\textwidth}{r *{4}{Y}}
\toprule
\textbf{Method} & \textbf{LBR Med} & \textbf{Panda} & \textbf{3D Gantry} & \textbf{duAro1} \\
\midrule
\text{SF-ResNet-E2E} & 0.00 & 0.00 & 0.00 & 0.00 \\
\text{SF-SFE-E2E} & 33.33 & 66.67 & \textbf{100.00} & 0.00 \\
\text{SF-ResNet-PT} & 33.33 & 66.67 & 66.67 & 0.00 \\
\text{SF-SFE-PT} & \textbf{66.67} & 66.67 & 33.33 & 0.00 \\
\midrule
\text{HHD-ResNet-E2E} & 0.00 & 0.00 & 33.33 & 0.00 \\
\text{HHD-SFE-E2E} & 33.33 & \textbf{100.00} & 0.00 & 0.00 \\
\text{HHD-ResNet-PT} & 33.33 & 0.00 & \textbf{100.00} & \textbf{33.33} \\
\text{HHD-SFE-PT} & \textbf{66.67} & 66.67 & \textbf{100.00} & \textbf{33.33} \\
\bottomrule
\end{tabularx}
\end{table*}

\subsection{Embodiment-Specific Results}

Finally, we present embodiment-specific breakdowns of our two most informative metrics: Fréchet distance (Table~\ref{tab:frechet}) and success rate (Table~\ref{tab:success}). Fréchet distance results align with our qualitative observations of trajectories, some of which are visualized in Fig. \ref{fig:rollouts}. For success rate, we exclude TAC which generated no successes. Our proposed \textbf{HHD-SFE-PT} model ranks among the most successful on every robot.

These results demonstrate that the learned policy generalizes well across embodiments with different morphologies and control interfaces, even though no student-robot data were used during training.

%% file: text/06_DL.tex
\section{Discussion and Limitations}\label{section:discussion}

This work demonstrates that vision-based tactile sensing can enable compliant, collaborative manipulation on position-controlled robots without joint-torque sensing or impedance control. By training on demonstrations from an impedance-enabled robot and leveraging a structured tactile representation—the Helmholtz-Hodge decomposition (HHD) of shear displacement fields—we achieve zero-shot policy transfer across four diverse robot embodiments. Our results show that tactile feedback can act as a practical proxy for force feedback, enabling policies that mimic the viscous control regime defined in Eq.~\ref{eq:viscous}.  

Among the tactile representations evaluated, HHD paired with our pre-trained scalar field encoder (SFE) achieved the highest overall success rates and lowest trajectory deviation. We hypothesize that this benefit arises because HHD de-aliases task-relevant information such as normal and shear forces, while filtering out environmental and sensor-specific artifacts present in raw tactile images. Nevertheless, other SFE-based policies performed competitively, suggesting the encoder itself provides a broadly useful low-dimensional tactile representation. Multimodal BC frameworks, such as Diffusion Policy \cite{chi2023}, could benefit from incorporating this encoder for contact-rich tasks.

We also evaluated six potential metrics and found that success rate and Fréchet distance were the most informative for assessing policy quality. Resistive force showed low discriminative power, as both failed and successful policies exhibited similar values, though reducing human-applied force remains an important goal.  

Several limitations remain. First, our experiments focus on tasks solvable entirely through human guidance; richer collaborative scenarios where robots balance human intent with autonomous objectives are not addressed here. Second, we restrict compliance to translational motion only, leaving rotational compliance for future work. Third, demonstrations and evaluations were conducted by expert operators, which may limit generalization; incorporating broader user studies would provide stronger evidence of usability.

Finally, while our method enables compliance-like behavior on position-controlled robots, resistive forces remain higher than under native impedance control, reflecting the inherent limitations of indirect force feedback. This high resistance may be an artifact of pseudo-velocity control via commanding $\Delta$-poses. Direct control of end-effector velocity should be implemented in future work.

Overall, our results establish vision-based tactile sensing and structured tactile representations as a promising proxy for force-torque feedback in enabling compliant, cross-embodiment collaborative manipulation, while highlighting avenues for improving policy robustness, tactile sensitivity, and broader task generalization.